\documentclass[conference, 10pt]{IEEEtran}
\usepackage[T1]{fontenc}
\usepackage[utf8]{inputenc}
\usepackage{mathptmx}

\usepackage{amsmath,amsfonts,amssymb}
\usepackage{bbm, comment}
\usepackage{gensymb, array}
\usepackage{latexsym}
\usepackage{graphicx}
\usepackage{multirow,rotating}
\usepackage{algpseudocode,algorithm}
\usepackage{epstopdf}
\usepackage[margin=0.5cm]{subcaption}
\usepackage[top = 19mm, left = 19mm, right = 19mm, bottom = 19mm]{geometry}
\usepackage{color}
\usepackage{tikz, pgfplots}
\newcommand{\argmin}{\operatornamewithlimits{argmin}}
\newcommand{\argmax}{\operatornamewithlimits{argmax}}

\allowdisplaybreaks

\title{Comparison of echo state network output layer classification methods on noisy data}

\author{\IEEEauthorblockN{Ashley A.\ Prater}
\IEEEauthorblockA{Air Force Research Laboratory\\
Information Directorate\\
Rome, NY 13441\\
Email: ashley.prater.3@us.af.mil}
}

\excludecomment{longversion}
\includecomment{shortversion}

\begin{document}
%\vspace{6mm}
\newgeometry{top = 25mm, left = 19mm, right = 19mm, bottom = 19mm}
\maketitle

\begin{abstract}
Echo state networks are a recently developed type of recurrent neural network where the internal layer is fixed with random weights, and only the output layer is trained on specific data.  Echo state networks are increasingly being used to process spatio-temporal data in real-world settings, including speech recognition, event detection, and robot control.  A strength of echo state networks is the simple method used to train the output layer - typically a collection of linear readout weights found using a least squares approach.  Although straightforward to train and having a low computational cost to use, this method may not yield acceptable accuracy performance on noisy data.

This study compares the performance of three echo state network output layer methods to perform classification on noisy data: using trained linear weights, using sparse trained linear weights, and using trained low-rank approximations of reservoir states.  The methods are investigated experimentally on both synthetic and natural datasets.  The experiments suggest that using regularized least squares to train linear output weights is superior on data with low noise, but using the low-rank approximations may significantly improve accuracy on datasets contaminated with higher noise levels.
\end{abstract}

%%%%%%%%%%%%%%%%%%%%%%%%%%%%%%%%%%%%%%%%%%%%%%%%%%%%%
%%
%%
\section{Introduction}

An Echo State Network (ESN) is a type of recurrent neural network, where the internal layer has randomly assigned connections and weights among the nodes.  ESNs, first introduced simultaneously in~\cite{jaeger} and in~\cite{maasrealtime} under the name \emph{Liquid State Machines} (LSM), have shown promise in performing classification and pattern prediction on spatio-temporal datasets.  ESNs and LSMs, collectively studied under the name \emph{Reservoir Computing}~\cite{verstraeten}, have been used in diverse applications such as speech recognition~\cite{maassNIPS, skowronski, verstraeten_words}, noise modeling~\cite{jaeger_haas}, event detection~\cite{hertzberg}, robot control~\cite{burgsteiner, joshi}, as well as chaotic time-series prediction~\cite{jaeger_lstm}.

Many variations of the behavior and topology of reservoirs have been studied, including both fully connected~\cite{jaeger} and sparse~\cite{luko2} random connections, or cyclical time-delay reservoirs~\cite{grigoryeva, paquotopto}, and the reservoir nodes themselves may be analog, leaky~\cite{jaegerleaky} or spiking~\cite{almassian}.  The hallmark of all reservoir designs is that the internal layer is fixed and is not trained for specific datasets or applications.  Non-reservoir recurrent-neural networks are typically trained by using a computationally expensive backpropagation method~\cite{werbos}.  By fixing the internal layer of a reservoir, all training occurs only at the output layer.  Commonly the output layer consists of a linear readout of the internal states, where the readout weights are trained using a regularized least squares approach.  Although this method is computationally fast, it may not produce the most accurate results, particularly in the presence of noisy data~\cite{praterarxiv, trefethen}.   Specialized ESNs have been proposed to handle noisy data, in particular embedded in support vector machines~\cite{shi} or in a Bayesian framework~\cite{li}, but done only in the context of time-series prediction tasks.

In contrast, this study compares the performance of comparatively basic reservoir output methods on noisy data for classification tasks.  This side-by-side study of the basic approaches will guide the development of better specialized methods for specific applications.  The three basic methods included within are:
\begin{enumerate}
	\item[A.] Using regularized least squares to train the linear readout weights.
	\item[B.] Using the Dantzig selector to train sparse, linear readout weights.
	\item[C.] Finding a low-rank approximation of the reservoir states for each class to determine membership of test data.
\end{enumerate}
Method A, using regularized least squares, was introduced in the seminal works, and still appears the most commonly in literature.  Its appeal is that it typically produces good results and is computationally fast.  However, it may be sensitive to reservoir parameters and noise.  Three variants of Method A are considered: (1) computing a new output weight matrix for each timestep, (2) computing a single output weight matrix based on the final timestep, and (3) computing a single global output weight matrix to fit all timesteps.  

Method B finds a sparse collection of readout weights.   This approach may be beneficial when the reservoir has significantly more nodes than necessary to fully distinguish the different classes of the input data.  Sparse readout weights were used with ESNs using computationally intense pruning methods~\cite{dutoit} and efficient elastic nets for time-series forecasting~\cite{bianchi}, but does not appear to be a widely adopted approach.  In this work, the sparse readout weights are obtained using the Dantzig selector~\cite{dantzig}, a regularized $\ell_1$ optimization problem similar to LASSO~\cite{lasso}.   It will be slightly slower to train than Method A, but will be equally fast in testing.  %Sparse readouts were obtained using computationally intense pruning methods in~\cite{dutoit} and using elastic nets for time-series forecasting in~\cite{bianchi}.  
%This work attains sparse readouts using the Dantzig selector~\cite{dantzig}, a regularized $\ell_1$ optimization problem similar to LASSO~\cite{lasso}.  

Finally, method C does not use a linear transformation of
\pagebreak
\restoregeometry
\noindent
the internal layer data, but forms collections of low-rank principal components of the reservoir responses to the input data of each class.  The reservoir responses of a new test signal are compared to these collections.  The signal is assigned to the class whose principal components best describe the variation in the reservoir response.  Using principal components to perform classification has been widely adopted in other contexts, but has only recently appeared for use with reservoirs~\cite{praterarxiv}.  This method should perform very well with noisy data, but may be slower to produce results.

All methods are compared theoretically and experimentally, with experimental results performed on synthetic and real-world data with artificial noise applied.  The results will show that in the noise-free case, the investigated methods typically yield similar classification accuracy.  For ESNs with a small number of hidden-layer nodes, the regularized least squares approaches in Method A tend to give the most accurate results, although accuracy degrades quickly as the noise level increases.  According to the experiments, better accuracy can  be achieved on noisy data by increasing the number of nodes and using the low-rank approximation technique as in Method C.  Only simple classification is performed; we do not employ ensemble classifiers.

This work is organized as follows:  Section~\ref{sec:models} describes the behavior of an ESN in detail, and discusses implementation of the three output methods explored in this work.  Section~\ref{sec:experimental results} reports on the performance of the three methods on two different datasets.  The article concludes with a summary of the experimental findings in Section~\ref{sec:conclusion}.   

The following notation is used.  The set of nonnegative integers is denoted by $\mathbb{N}_0$.  Let $e_k$ denote a vector, of length clear from the context, of all zeros except for a "1" in the $k^\text{th}$ entry.  The $\ell_p$ norm of a matrix is denoted by $\|\cdot\|_p$, for $1\leq p \leq \infty$.  The normal distribution with mean $\mu$ and standard deviation $\sigma$ is given by $N(\mu,\sigma)$, and the uniform distribution on the interval $[a,b]$ is given by $U(a,b)$. The transpose of a matrix $X$ is denoted by $X^\top$.

\section{ESN Models and Output Classifiers}\label{sec:models}

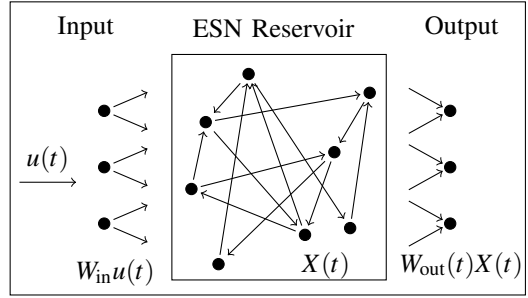
\begin{figure}[bthp]
\centering
\begin{tikzpicture}
	% Input
	\node at (-1,3.35) {Input};
	\node (In1) at (-0.75,0.75) {};
	\node (In2) at (-0.75,1.5) {};
	\node (In3) at (-0.75,2.25) {};

	\node (In1b) at (-.05,.45) {};
	\node (In1t) at (-.05,1.05) {};
	\node (In2b) at (-.05,1.2) {};
	\node (In2t) at (-.05,1.8) {};
	\node (In3b) at (-.05,1.95) {};
	\node (In3t) at (-.05,2.55) {}; 

	\draw [black,fill] (In1) circle (0.075);
	\draw [black,fill] (In2) circle (0.075);
	\draw [black,fill] (In3) circle (0.075);

	\draw [->] (In1) edge (In1b);
	\draw [->] (In1) edge (In1t);
	\draw [->] (In2) edge (In2b);
	\draw [->] (In2) edge (In2t);
	\draw [->] (In3) edge (In3b);
	\draw [->] (In3) edge (In3t);
	
	\node (u) at (-1.5,1.6) {$u(t)$};
	\node (ul) at (-2,1.3) {};
	\node (ur) at (-1,1.3) {};
	\draw [->] (ul) edge (ur);

	\node at (-0.6,0.1) {$W_\text{in}u(t)$};

	% Output
	\node at (4,3.35) {Output};
	\node (O1) at (4,1.5) {};

	\draw [->] (3.3,2.55) -- (3.75,2.3);
	\draw [->] (3.3,1.95) -- (3.75,2.2);

	\draw [->] (3.3,1.8) -- (3.75,1.55);
	\draw [->] (3.3,1.2) -- (3.75,1.45);

	\draw [->] (3.3,1.05) -- (3.75,0.8);
	\draw [->] (3.3,0.45) -- (3.75,0.7);

	\draw [black,fill] (3.85,2.25)  circle (0.075);
	\draw [black,fill] (3.85,1.5)  circle (0.075);
	\draw [black,fill] (3.85,0.75)  circle (0.075);

	\node at (4,0.2) {$W_\text{out}(t)X(t)$};

	\draw [black, thin] (-2,-.2) rectangle (4.85,3.7);

	% Reservoir nodes
	\draw [black,thin] (0.15,0) rectangle (3,3);
	\node at (1.5,3.35) {ESN Reservoir};

	\node (A) at (0.41, 1.21) {};
	\node (B) at (0.60, 2.10) {};
	\node (C) at (1.17, 2.74) {};
	\node (D) at (1.92, 0.60) {};
	\node (E) at (0.77, 0.21) {};
	\node (F) at (2.31, 1.70) {};
	\node (G) at (2.52, 0.69) {};
	\node (H) at (2.78, 2.49) {};

	\node at (2.2,.2) {$X(t)$};

	%	Draw circles at nodes
	\draw [black, fill] (A) circle (0.075);
	\draw [black, fill] (B) circle (0.075);
	\draw [black, fill] (C) circle (0.075);
	\draw [black, fill] (D) circle (0.075);
	\draw [black, fill] (E) circle (0.075);
	\draw [black, fill] (F) circle (0.075);
	\draw [black, fill] (G) circle (0.075);
	\draw [black, fill] (H) circle (0.075);

	% Connections among nodes
	% 1-2 and 1-6
	\draw [->] (A) edge (B);
	\draw [->] (A) edge (F);
	\draw [->] (B) edge (D);
	\draw [->] (B) edge (H);
	\draw [->] (C) edge (B);
	\draw [->] (C) edge (G);
	\draw [->] (D) edge (A);
	\draw [->] (D) edge (C);	
	\draw [->] (E) edge (C);
	\draw [->] (F) edge (D);
	\draw [->] (F) edge (E);
	\draw [->] (G) edge (H);
	\draw [->] (H) edge (F);
\end{tikzpicture}
\caption{A representation of an ESN with sparse connections among the internal nodes.}
\label{fig:ESN model}
\end{figure}

The behavior of ESNs is described in detail here.  Let $u$ be an input pattern, and let $u(t)$ denote its value at time $t\geq 0$.  Suppose $u$ is discretized and parameterized so $t\in\mathbb{N}_0$.  Suppose $u$ is fed into an random ESN with $N$ nodes, possibly leaky, with no output feedback.   Then the internal-layer \emph{reservoir state} at time $t$ is given by the vector $X(t)\in\mathbb{R}^N$, whose value can be determined by
\begin{align}
	X(t+1) = &(1-\alpha)X(t) + \label{eq:esn} \\
		&f\left( \rho W_\text{res}X(t) + \gamma W_\text{in}u(t) + W_\text{fb}\hat{y}(t) \right) \nonumber \\
	\hat{y}(t+1) = &g(W_\text{out} S(t)), \label{eq:output}
\end{align}
Here, the ESN model appearing in~\cite{jaegerleaky} is followed.  A representation of this design of an ESN is shown in Figure~\ref{fig:ESN model}.  Throughout this article we set $g(x) = x$ and $W_\text{fb} = \bf {0}$, since nonzero feedback is inconsistent with the third output classification method in this study.  $S(t)$ appearing in~\eqref{eq:output} may be chosen as either $X(t)$ or a concatenation such as $[X(t); u(t)]$. For notational convenience, $X$ is used throughout this section, but the results hold similarly for $S$ with appropriate changes to matrix dimensions.   

The parameters $\rho$ and $\gamma$ are the feedback strength and input gain, respectively.  The leaking rate is given by $\alpha \in [0,1]$. The input weights $W_\text{in}$ are a fixed $N\times L$ matrix, where $L$ is the dimension of the input patterns at each time step.  The reservoir weights $W_\text{res}$ are a fixed $N\times N$ matrix.  The output weights $W_\text{out}$ are learned $K\times N$ matrices, where $K$ is the number of classes of the input data.

%%%%%%%%%%%%%%%%%%%%%%%%%%%%%%%%%%%%%%%%%%%%%%%%%%%%%%%%%%%%%%%%%%%%%%
\subsection{Output weights trained by regularized least squares}\label{sec:RLS}

Traditionally in reservoir computing literature, input signals are classified by using a linear combination of the reservoir states, where the readout weights are trained using a least squares method~\cite{jaeger, luko2, goudarzi, luko}.  Three variations of the trained linear output weights are explored: temporal pointwise weights, single weights trained against the final timestep, and global weights trained against all timesteps.  To determine the readout weights using any of these variations, first let $\{u_j\}_{j=1}^J$ be a collection of distinct training patterns, each belonging to one of $K$ classes.  Compute $X_j(t) \in\mathbb{R}^N$ the reservoir states of each $u_j$ at time $t$ according to Equation~\eqref{eq:esn}.  Collect these states into matrices $X_\text{tr}(t) \in \mathbb{R}^{N\times J}$ by concatenating columns.  Define a matrix $y \in\mathbb{R}^{K\times J}$ indicating class membership of the training set.  Say $y(k,j) = 1$ if $u_j$ is in the $k^\text{th}$ class and either $0$ (or possibly $-1$) otherwise.

%%%%%%%%%%%%%%%%%%%%%%%%%%%%%%%%%%%
\vspace{.05 in}
\emph{ A1. Pointwise output weights} 
\vspace{0.05 in}

The input patterns may display large variation with respect to the time variable $t$, even within the same class.  In this case, a new output weight matrix should be computed for each $t$.  That is, the output weight matrices $W_\text{out}^\text{pt}(t) \in \mathbb{R}^{K\times N}$ are determined so that $W_\text{out}^\text{pt}(t) X_\text{tr}(t) \approx y$.  This can be done using least squares, but the result may be very sensitive to small purturbations~\cite{trefethen}.  To obtain a more robust result, the regularized least squares method~\eqref{eq:RLS} is employed: For each $t$, let $W^\text{pt}_\text{out}(t)$ be obtained via
\begin{equation}\label{eq:RLS}
	W_\text{out}^\text{pt}(t) = \argmin_{W\in\mathbb{R}^{K\times N}} \left\| y - WX_\text{tr}(t) \right\|_2^2 + \lambda \left\| W\right\|_2^2.
\end{equation}

The value $\lambda$ appearing in Equation~\eqref{eq:RLS} is the regularization parameter, and helps to minimize overfitting of the training data.  A larger $\lambda$ will force dampening of the coefficients in $W_\text{out}^\text{pt}$.  A smaller $\lambda$ forces the approximation $y\approx W_\text{out}^\text{pt}X$ to be closer.  An appropriate $\lambda$ must be chosen to balance these two goals, and may be user-specified or determined using a cross-validation technique as in~\cite{shi, li}.  Equation~\eqref{eq:RLS} can be solved explicitly by 
\begin{equation*}
	W_\text{out}^\text{pt}(t) = yX^\top_\text{tr}(t) \left( X_\text{tr}(t)X^\top_\text{tr}(t) + \lambda I_{N\times N}\right)^{-1}.
\end{equation*}

After determining the collection of output weights, a newly encountered pattern is classified depending on how well the weights $W_\text{out}^\text{pt}$ project reservoir states onto columns of $y$.  That is, let $u$ be the test pattern, and $X_u(t)$ be its reservoir states determined by Equation~\eqref{eq:esn}.  If $u$ belongs to the $k^\text{th}$ class, then theoretically $W_\text{out}(t)X_u(t)\approx e_k$ for most $t$.  More precisely, let 
\begin{equation}\label{eq:z}
	z = \sum_{t} W^\text{pt}_\text{out}(t)X_u(t).
\end{equation}
Then $u$ is predicted to be in the $k^\text{th}$ class if $k = \argmax{z}$.

This approach should work well on datasets with patterns that display similar intra-class temporal behavior, but has a somewhat high computational cost.

%%%%%%%%%%%%%%
\vspace{0.05 in}
\emph{ A2. Final output weights}
\vspace{0.05 in}

In this approach, a single weight matrix is found based on the final temporal state of the reservoir.  Suppose $X_\text{tr}^\text{end}\in\mathbb{R}^{N\times J}$ is the collection of final reservoir states for all $J$ patterns in the training set.  The weight matrix $W_\text{out}^\text{end} \in \mathbb{R}^{K\times N}$ is found so that $W_\text{out}^\text{end}X_\text{end} \approx y$.  Here, the matrix is found using the following regularized optimization problem:
\begin{equation}\label{eq:end}
	W_\text{out}^\text{end} = \argmin_{W\in\mathbb{R}^{K\times N}}\left\{ \left\|y - WX_\text{tr}^\text{end}\right\|_2^2 + \lambda\|W\|_2^2\right\}
\end{equation}

A newly encountered pattern $u$ is classified based on its final reservoir state $X_u^\text{end}$.  Let $z = W_\text{out}^\text{end}X_u^\text{end}$.  Then $u$ is predicted to be in the $k^\text{th}$ class if $k = \argmax{z}$.  

This approach has a lower computational cost than the pointwise approach in (A1), and should work well if the final reservoir states are well separated for different classes.  However, since classification is based on just a single point in time, accuracy may degrade if noise propagates through the reservoir and sufficiently contaminates the final reservoir states.

%%%%%%%%%%%%%%%%%%%%%%%%%%%%
\vspace{0.05 in}
\emph{ A3. Global output weights}
\vspace{0.05 in}

In this approach, a single weight matrix is found based upon all of the reservoir responses.  The weight matrix $W_\text{out}^\text{glob} \in \mathbb{R}^{K\times N}$ is found so that $X_\text{out}^\text{glob} X_\text{tr}(t) \approx y$ \emph{for all} $t$.   Here, the matrix is found using regularized least squares:
\begin{equation}\label{eq:global}
	W_\text{out}^\text{glob} = \argmin_{W\in\mathbb{R}^{K\times N}} \left\{ \sum_t \left\|y - WX_\text{tr}(t) \right\|_2^2 + \lambda \|W\|_2^2\right\}.
\end{equation}

A newly encountered pattern $u$ is classified similarly to the previous two approaches.  Let $X_u(t)$ be the reservoir state for the input $u$ at time $t$, and let 
\begin{equation*}
	z = \sum_t W_\text{out}^\text{glob} X_u(t).
\end{equation*}
Then $u$ is predicted to be in the $k^\text{th}$ class if $k = \argmax z$.

In essence, this approach separates the average reservoir state over time of the classes.  It should perform well on datasets with well separated class means and low intra-class variation.

%%%%%%%%%%%%%%%%%%%%%%%%%%%%%%%%%%%%%%%%%%%%%%%%
\subsection{Sparse trained output weights}

This method produces sparse linear output weights.  The output weights determined by methods A1, A2 and A3 will in general be dense, with every reservoir node making a nonzero contribution to the output layer.  However, it is reasonable to assume that only a subset of the reservoir nodes can be used for performing classification, especially if the number of nodes in the reservoir is very large.  To this end, a sparse matrix recovery method may be used, such as LASSO~\cite{lasso} or the Dantzig selector~\cite{dantzig}, to determine sparse output weights $W_\text{out}^\text{sp}(t)$.  

We employ the Dantzig selector, which works as follows.  Let $X_\text{tr}(t)$ be the collection of reservoir nodes of the training set, and let $y$ be the matrix indicating class membership.  For each $t$, compute
\begin{equation}\label{eq:dan}
	W_\text{out}^\text{sp}(t) = \argmin_{W\in\mathbb{R}^{K\times N}} \left\{ \left\| \left(WX_\text{tr}(t) - y\right)X(t)^\top D^{-1}\right\|_\infty + \lambda\|W\|_1\right\}.
\end{equation}

The value $\lambda>0$ is a fixed regularization parameter and $D$ is a diagonal scaling matrix with entries equal to the $\ell_2$ norm of the rows of $X(t)$.  Similar to Equation~\eqref{eq:RLS}, a larger $\lambda$ forces $W^\text{sp}_\text{out}$ to be sparser with small nonzero coefficients, while a smaller $\lambda$ makes the approximation $W^\text{sp}_\text{out}X_\text{tr} \approx y$ better.  The parameter $\lambda$ must be chosen to balance accuracy and sparsity.  Equations~\eqref{eq:RLS} and~\eqref{eq:dan} are similar in that they both have terms to encourage close approximations, however, the regularization term of Equation~\eqref{eq:dan} controls the overall sparsity of $W_\text{out}$, rather than the regularization term of~\eqref{eq:RLS} that tends to make the entries of $W_\text{out}$ more uniform in magnitude.  Equation~\eqref{eq:dan} can be solved using the algorithms appearing in~\cite{prater_dantzig} or~\cite{lupong}.

The test phase for this method follows that in Subsection~II-A1.

%%%%%%%%%%%%%%%%%%%%%%%%%%%%%%%%%%%%%%%%%%%%%%%%
\subsection{Principal components of reservoir states}

This method performs classification based on the principal components of the reservoir states.  This approach is motivated by the fact that often inputs from the same class have similar underlying reservoir responses.  Since ESNs operate at the ``edge of chaos"~\cite{leg}, perturbations of inputs within the same class may lead to noisy discrepancies in the corresponding reservoir states, however, a principal component based method will identify any strong unifying underlying reservoir response pattern if it exists.

The method employed below is a variant of that found in~\cite{praterarxiv}. 
In the training phase, for each time $t$, create $K$ matrices $X_k(t)$ each one equaling the collection of reservoir states from inputs in the $k^\text{th}$ class of the training set, determined using Equation~\eqref{eq:esn}.  For some fixed positive integer $R$, let ${U_k(t)\in\mathbb{R}^{N\times R}}$ be the collection of the first $R$ principal components of $X_k(t)$.

In the testing phase, let $u$ be a newly encountered input signal and $X(t)$ its reservoir responses.  If $u$ belongs to the $k^\text{th}$ class, then the columns of $U_k$ should capture the behavior of $X$ well at each $t$.  Therefore the class membership of $u$ may be predicted as follows. Let $z$ be a $K$-vector with
\begin{equation}\label{eq:pca}
	z(k) =  \sum_t \left\|\left( U_k(t) U_k(t)^\top - I_{N\times N}\right) X(t)\right\|.
\end{equation}
Say $u$ is in the $k^\text{th}$ class if $k = \argmin z$.

Since noise is a high-frequency component overlaying the true signal, by considering the low-rank approximations, this method should adapt to noisy data well.

\section{Experimental Results}\label{sec:experimental results}

In this section, the output classification methods are compared experimentally.  Two datasets are used: A synthetic dataset with signals randomly alternating between sine waves and square waves, and a real-world dataset consisting of samples of speakers uttering the Japanese vowel `ae'. 

The sine/square wave dataset has been used in reservoir computing experiments, notably in~\cite{paquotopto} and~\cite{zhang} for study with photonic reservoirs.  The Japanese vowel dataset (JV) was used in~\cite{jaegerleaky} with ensemble classifiers built from several small leaky ESNs.

The output layer methods are explored using these datasets in a noise-free case, as well as when the data are modified by applying gaussian noise $N(0,\sigma)$ to the input for various levels $\sigma$.  The reservoirs are initialized as in Equation~\eqref{eq:esn} for various $N$.  For each pair $(N,\sigma)$, a number of simulations are performed, with any randomizations in the matrices $W_\text{in}$ and $W_\text{res}$, the signals $u$,  or in the application of noise redetermined for each simulation.  

For both datasets, two regularization parameters are used with the output layer method A1 of pointwise regularized least squares, $\lambda = 10^{-4}$ and $\lambda = 10^{-10}$, which are labeled as `(A1) $\lambda = 10^{-4}$' and `(A1) $\lambda = 10^{-10}$' respectively in the tables and figures of results.  For the output layer methods A2, A3, and B, the regularization parameter $\lambda = 10^{-4}$ is used.  These results are labeled as `(A2) Endpoints', `(A3) Global', and `(B) Sparse' respectively.  Finally, the results using method C using low-rank approximations are labeled as `(C) PCA'.  

A small number of regularization parameters are used, providing a snapshot comparison of their performance for various noise levels.  The results show that a larger regularization parameter tends to yield higher classification accuracy for data with a higher level of noise contamination.%  In practice, after an output method is selected, one may wish to determine the appropriate regularization parameter through a cross-validation technique.

The sine/square wave experiments are performed in MATLAB 2015b on a PC at 2.40 GHz and 16 GB RAM.  The JV experiments are performed in MATLAB 2016a on the Condor Cluster hosted at the RIT DOD HPC Affiliated Resource Center.

%%%%%%%%%%%%%%%%%%%%%%%%%%%%%%%%%%%%%%%%
\subsection{Sine vs.\ Square Wave}
\begin{figure}[htp]
	\centering
	\includegraphics{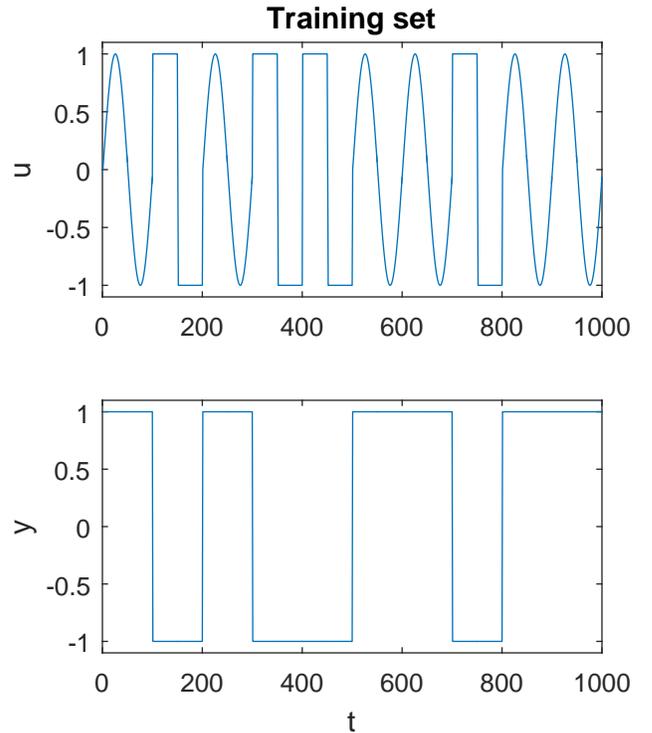}
	\caption{A sample of the training set in the sine/square wave example (top) with the associated indicator function (bottom).}
	\label{fig:ss training}
\end{figure}

In this collection of experiments, the training dataset is a signal $u$ with randomly placed sine and square wave segments, along with an indicator function $y$, where 
\begin{equation}\label{eq:ss indicator}
	y(t) = \begin{cases}
		1, &\text{ if $u$ is a sine wave at time $t$}, \\
		-1, &\text{ if $u$ is a square wave at time $t$}.
		\end{cases}
\end{equation}
A sample of a typical training set is shown in Figure~\ref{fig:ss training}.  The individual sine and square wave segments have the same period.  The reservoir responses are found as in Equation~\eqref{eq:esn}, with both $W_\text{in}$ and $W_\text{res}$ randomly choosing entries from $U(-1,1)$, but $W_\text{res}$ scaled so that its spectral radius is less than $1$.  The parameters are chosen as $\rho = 0.8$, $\gamma = 1.5$, and $\alpha = 1$.  The reservoir size $N$ is selected from $\{10, 25, 50, 75, 100\}$. 

The test set is generated by randomly combining the sine and square wave segments, then adding noise $\varepsilon \sim N(0,\sigma)$, for $\sigma \in \{0.00, 0.01, 0.02, 0.05, 0.10, 0.15, 0.20, 0.25, 0.30\}$ to the inputs of the test set.  No artificial noise is applied to the training set.  %For each pair $(N,\sigma)$ a new test set is generated and classified $500$ times.  

\begin{table*}
	% Data commented out is for 'clean' training set.  The black is for the noisy training and test sets.
	\centering
	\begin{tabular}{c|c||rr|rr|rr|rr|rr|rr}
	\multicolumn{2}{c|}{}	&\multicolumn{2}{c|}{(A1) $10^{-4}$} &\multicolumn{2}{c|}{(A1) $10^{-10}$} 	&\multicolumn{2}{c|}{(A2) Endpoints} 	&\multicolumn{2}{c|}{(A3) Global} &\multicolumn{2}{c|}{(B) Sparse} &\multicolumn{2}{c}{(C) PCA}\\
	$N$	&$\sigma$		&mean 	&std		&mean  &std 	&mean		&std		&mean  &std 	&mean	&std &mean &std \\
	\hline
	\multirow{8}{*}{10}
%	Done on HPC, 28 Nov 2016
%	SinSquareTrial4.m
&0.00  &98.48 &0.32      &98.48 &0.32     &{\bf \color{red}100.00} &0.00      &50.09 &0.27     &87.87 &3.20      &98.48 &0.32 \\
&0.01  &92.67 &0.63      &92.46 &0.61     &{\bf \color{red}99.92} &0.57      &50.25 &0.22     &87.53 &3.62        &94.05 &0.48  \\
&0.02  &89.93 &0.67      &89.65 &0.78     &{\bf \color{red}96.88} &5.50      &50.14 &0.38     &87.24 &2.52        &91.75 &0.43  \\
&0.05  &82.43 &1.22      &82.10 &1.30     &76.40 &20.40      &50.21 &0.71     &80.98 &1.93      &{\bf \color{red}87.15} &0.74  \\
&0.10  &72.43 &1.57      &72.18 &1.54     &63.52 &16.28      &50.25 &0.83     &71.69 &1.64      &{\bf \color{red}81.03} &0.84  \\
&0.15  &66.81 &1.91      &66.63 &1.94     &59.28 &19.49      &50.20 &0.66     &66.26 &1.70      &{\bf \color{red}75.84} &1.40  \\
&0.20  &62.04 &1.70      &61.86 &1.76     &60.32 &17.52      &50.05 &0.87     &61.50 &1.76      &{\bf \color{red}71.98} &1.28 \\
&0.25  &58.47 &1.70      &58.34 &1.72     &58.40 &17.33      &49.92 &0.82     &58.20 &1.63      &{\bf \color{red}68.51} &1.09 \\
&0.30  &55.82 &1.78      &55.59 &1.92     &55.12 &16.17      &50.00 &0.89     &55.39 &1.52      &{\bf \color{red}65.71} &1.58 \\
	\hline
\multirow{9}{*}{25}
%	Done on HPC, 28 Nov 2016
%	SinSquareTrial4.m, comparing pcatol and RPCA
&0.00  &98.58 &0.23      &98.58 &0.23     &{\bf \color{red}100.00} &0.00      &50.06 &0.17     &94.00 &1.21      &98.58 &0.23 \\
&0.01  &92.73 &0.61      &92.57 &0.63     &{\bf \color{red}99.52} &2.87      &50.24 &0.23     &92.41 &0.93       &94.02 &0.42  \\
&0.02  &89.74 &0.85      &89.57 &0.83     &{\bf \color{red}95.04} &9.98      &50.37 &0.39     &89.29 &1.01       &91.66 &0.50  \\
&0.05  &82.32 &1.00      &82.21 &1.03     &77.76 &17.35      &50.26 &0.60     &82.25 &1.02      &{\bf \color{red}87.08} &0.71  \\
&0.10  &72.33 &1.78      &72.17 &1.81     &64.40 &19.08      &50.27 &0.71     &72.11 &1.61      &{\bf \color{red}81.29} &0.94  \\
&0.15  &66.60 &1.81      &66.50 &1.80     &61.60 &20.24      &50.03 &0.76     &66.51 &1.60      &{\bf \color{red}76.26} &0.95   \\
&0.20  &62.08 &1.86      &61.99 &1.89     &58.48 &15.12      &49.95 &0.75     &62.01 &1.67      &{\bf \color{red}72.24} &1.22  \\
&0.25  &57.79 &1.69      &57.70 &1.73     &55.36 &17.51      &50.14 &0.85     &57.66 &1.61      &{\bf \color{red}68.89} &1.54   \\
&0.30  &55.53 &1.79      &55.47 &1.84     &56.32 &16.24      &50.07 &0.77     &55.50 &1.71      &{\bf \color{red}65.56} &1.90 \\
	\hline
\multirow{9}{*}{50}
%	Done on HPC, 28 Nov 2016
%	SinSquareTrial4.m, comparing pcatol and RPCA
&0.00  &98.51 &0.34      &98.51 &0.34     &{\bf \color{red}100.00} &0.00      &50.01 &0.22     &95.99 &0.95      &98.51 &0.34  \\
&0.01  &92.56 &0.67      &92.40 &0.69     &{\bf \color{red}99.92} &0.57      &50.19 &0.22     &92.36 &0.82        &94.06 &0.46  \\
&0.02  &89.78 &0.83      &89.65 &0.84     &{\bf \color{red}96.96} &5.91      &50.13 &0.49     &89.61 &0.86        &91.83 &0.43  \\
&0.05  &81.94 &1.09      &81.80 &1.09     &70.96 &22.67      &50.14 &0.63     &81.77 &1.09      &{\bf \color{red}87.16} &0.54  \\
&0.10  &72.89 &1.60      &72.85 &1.62     &63.68 &21.65      &50.03 &0.68     &72.81 &1.50      &{\bf \color{red}81.40} &0.80  \\
&0.15  &66.21 &2.06      &66.13 &2.07     &59.44 &20.87      &50.04 &0.96     &66.13 &1.92      &{\bf \color{red}76.18} &1.11   \\
&0.20  &61.99 &1.80      &61.97 &1.77     &57.20 &16.99      &49.90 &0.90     &61.92 &1.65      &{\bf \color{red}72.52} &1.40   \\
&0.25  &58.38 &2.01      &58.34 &2.02     &57.92 &19.72      &50.00 &1.03     &58.36 &1.98      &{\bf \color{red}68.64} &1.67   \\
&0.30  &55.90 &1.78      &55.85 &1.79     &56.08 &18.28      &50.01 &0.99     &55.86 &1.74      &{\bf \color{red}66.19} &1.77  \\
	\hline
\multirow{9}{*}{75}
%	Done on HPC, 28 Nov 2016
%	SinSquareTrial4.m, comparing pcatol and RPCA
&0.00  &98.50 &0.33      &98.50 &0.33     &{\bf \color{red}100.00} &0.00      &50.00 &0.22     &96.01 &0.89      &98.50 &0.33 \\
&0.01  &92.61 &0.55      &92.46 &0.59     &{\bf \color{red}99.44} &1.81      &50.25 &0.23     &92.34 &0.64        &93.97 &0.45  \\
&0.02  &89.66 &0.66      &89.59 &0.65     &{\bf \color{red}95.20} &9.28      &50.22 &0.46     &89.51 &0.74        &91.72 &0.50  \\
&0.05  &82.50 &1.07      &82.44 &1.08     &74.48 &21.53      &49.99 &0.75     &82.40 &0.99      &{\bf \color{red}87.00} &0.70  \\
&0.10  &72.79 &1.89      &72.76 &1.89     &59.12 &20.92      &50.11 &0.92     &72.75 &1.85      &{\bf \color{red}81.24} &0.99  \\
&0.15  &66.69 &1.54      &66.66 &1.57     &56.40 &18.06      &50.09 &1.02     &66.62 &1.52      &{\bf \color{red}76.38} &0.93   \\
&0.20  &61.45 &1.88      &61.42 &1.87     &58.96 &19.46      &50.07 &0.83     &61.46 &1.83      &{\bf \color{red}72.04} &1.18   \\
&0.25  &58.53 &2.09      &58.47 &2.10     &59.92 &16.77      &50.04 &0.93     &58.50 &2.07      &{\bf \color{red}68.89} &1.80   \\
&0.30  &56.05 &1.70      &56.00 &1.76     &52.24 &17.75      &50.28 &0.87     &56.01 &1.78      &{\bf \color{red}65.55} &1.77  \\
\hline
\multirow{9}{*}{100}
%	Done on HPC, 28 Nov 2016
%	SinSquareTrial4.m, comparing pcatol and RPCA
&0.00  &98.58 &0.29      &98.58 &0.29     &{\bf \color{red}100.00} &0.00      &50.05 &0.20     &96.20 &0.78      &98.58 &0.29  \\
&0.01  &92.52 &0.70      &92.40 &0.71     &{\bf \color{red}99.44} &2.43      &50.21 &0.22     &92.39 &0.71       &94.07 &0.38   \\
&0.02  &89.79 &0.83      &89.71 &0.82     &{\bf \color{red}94.72} &9.19      &50.20 &0.60     &89.69 &0.82      &91.74 &0.41   \\
&0.05  &82.32 &1.11      &82.28 &1.13     &71.44 &21.14      &50.04 &0.56     &82.30 &1.07      &{\bf \color{red}87.07} &0.60  \\
&0.10  &72.60 &1.47      &72.59 &1.44     &63.20 &21.59      &50.04 &0.89     &72.52 &1.38      &{\bf \color{red}81.13} &0.96   \\
&0.15  &66.62 &2.06      &66.61 &2.07     &60.72 &15.62      &50.30 &0.88     &66.58 &2.04      &{\bf \color{red}76.31} &1.01   \\
&0.20  &61.85 &2.01      &61.84 &2.01     &57.04 &18.03      &50.03 &0.80     &61.84 &1.97      &{\bf \color{red}72.39} &0.88   \\
&0.25  &58.75 &1.56      &58.73 &1.56     &52.48 &15.33      &49.99 &0.87     &58.73 &1.51      &{\bf \color{red}69.01} &1.49   \\
&0.30  &55.90 &1.78      &55.86 &1.77     &57.52 &15.22      &50.11 &1.03     &55.85 &1.76      &{\bf \color{red}66.07} &1.86   \\
\hline
\end{tabular}
\caption{Means and standard deviations of the percent classification accuracy over 50 experiments on the noisy sine/square wave test dataset using various classification methods.  For each reservoir size $N$ and noise level $\sigma$, the result of the best performing method is highlighted in red boldface font.}
\label{tab:ss accuracy}
\end{table*}

\begin{figure*}
	\centering
	\includegraphics{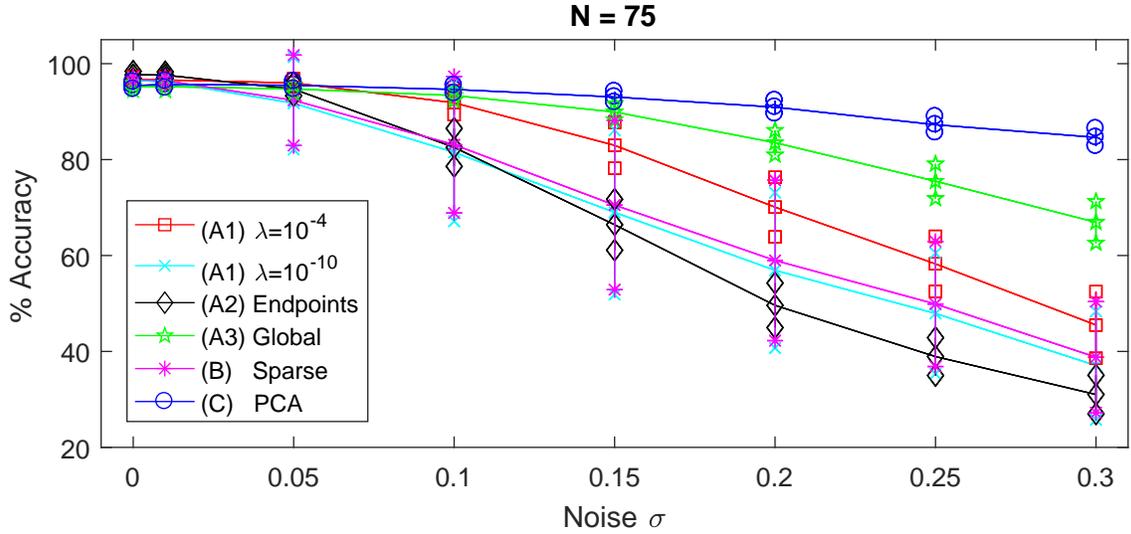}
\caption{Percent classification accuracy of the noisy Japanese vowel test dataset for various noises $\sigma$ with reservoir size fixed at $N=75$. The solid lines give the mean over all 100 simulations for each noise level $\sigma$, and the vertical lines represent one standard deviation from the mean. }
\label{fig:JPAcc}
\end{figure*}

\begin{figure*}
	\centering
	\includegraphics[width=0.48\textwidth]{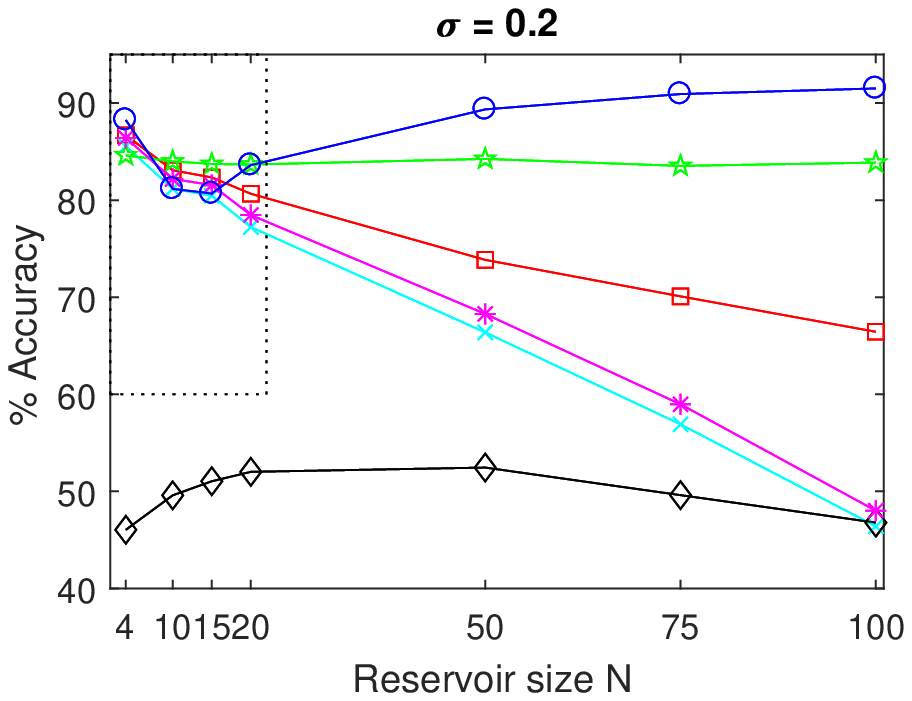}
	\includegraphics[width=0.48\textwidth]{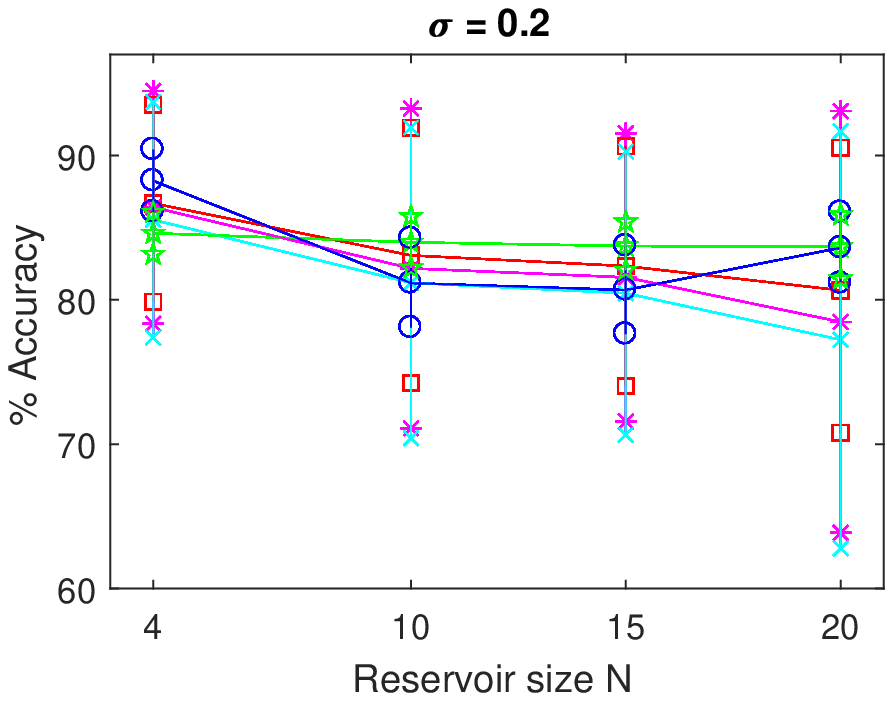}
\caption{Percent classification accuracy of the noisy Japanese vowel test dataset for various reservoir sizes $N$ with noise fixed at $\sigma = 0.2$.  The figure on the left displays the means for the various methods.  The figure on the right displays mean and standard deviation details for the portion inside the dotted box on the left.  The colors and marker styles match those in Figure~\ref{fig:JPAcc}.}
\label{fig:JV Accuracy}
\end{figure*}

%\begin{figure*}
%	\centering
%	\begin{subfigure}{0.48\textwidth}
%		\includegraphics[0.98\textwidth]{Figure6a.eps}
%		\caption{Means of percent classification accuracies.}
%		\label{fig:6a}
%	\end{subfigure}
%	\begin{subfigure}{0.48\textwidth}
%		\includegraphics[0.98\textwidth]{Figure6b.eps}
%		\caption{Means and standard deviations of percent classification accuracies for `boxed' area of Figure~\ref{fig:6a}.}
%		\label{fig:6b}
%	\end{subfigure}
%\caption{Percent classification accuracy of the noisy Japanese vowel test dataset for various reservoir sizes $N$ with noise fixed at $\sigma = 0.2$.  The figure on the left displays the means for the various methods.  The figure on the right displays mean and standard deviation details for the portion inside the dotted box on the left.  The colors and marker styles match those in Figure~\ref{fig:JPAcc}.}
%\label{fig:JV Accuracy}
%\end{figure*}

Table~\ref{tab:ss accuracy} displays the mean and standard deviation of the test accuracy for the 50 simulations for each pair $(N,\sigma)$.  At the noise-free level, methods A1, A2, A3, and C perform similarly, with the endpoint method A3 giving the best results.  This is expected, due to the large discrepancy in the final value of the inputs for each class ($-1$ for square waves vs $0$ for sines).  As the noise $\sigma$ increases, the low-rank method C outperforms the others as expected. The performance of the sparse method B improves as the reservoir size increases, for then it has more dimensions from which to choose the few most significant nodes contributing to classification.  The performance of the other methods do not depend on $N$, probably due to the simple behavior of the input patterns.  Finally, note that the single global output weight method B performs poorly.  This is likely explained by the fact that both the sine and square waves have an average value of 0.

\subsection{Japanese Vowels}
In this collection of experiments, speaker identification is performed using the Japanese vowel dataset.  
The dataset originally appeared in~\cite{kudo} and was obtained via~\cite{lichman}.  JV is a temporal set with samples recorded from nine male speakers saying the Japanese vowel `ae'. Each sample forms an $m\times 12$ real-valued matrix, where $m$ is an integer between 7 and 29 (inclusive) depending on the duration of the utterance, and 12 is the number of cepstrum coefficient features observed for each speaker.  The dataset includes 640 samples; of which a fixed selection of 270 are used for training (with 30 utterances by each of the nine speakers), and the remaining are used for testing (24-88 utterances by each of the nine speakers).   The test dataset was modified by adding gaussian white noise ${\sim~N(0,\sigma)}$, with $\sigma \in \{0.00, 0.01, 0.05, 0.10, \ldots, 0.30\}$.

Results were obtained using a simulated ESN, with ${L=14}$ input nodes accounting for the 12 cepstrum coefficients plus two bias terms, ${K=9}$ output nodes representing the 9 classes (one for each speaker), and N reservoir nodes with $N\in\{4, 10, 15, 20, 50, 75, 100\}$.  The ESN parameters were fixed at ${\alpha = 0.2}, {\rho = 0.2}, {\gamma = 1.5}$, and  ${f = \tanh}$, chosen to match those appearing in~\cite{jaegerleaky}.

Table~\ref{tab:jv accuracy} displays the mean and standard deviation of the test classification accuracy over 100 simulations for each pair $(N,\sigma)$.  For low noise levels, the methods perform similarly, especially for larger $N$, with the endpoint method A2 outperforming the rest.  As expected, the endpoint method A2 appears to be the least tolerant to noise.  The single global weight method A3 is somewhat tolerant to noise corruption, but the low-rank method C is shown to be the most robust to noise.  The table suggests that to achieve the best classification accuracy on low noise data, one should use the endpoint method A2 with a moderate reservoir size, and to achieve the best accuracy on higher noise data, one should use the low-rank approach C with a large reservoir size.

\begin{table*}
	\centering
	\begin{tabular}{c|c||cc|cc|cc|cc|cc|cc}
	\multicolumn{2}{c|}{}	&\multicolumn{2}{c|}{(A1) $10^{-4}$} &\multicolumn{2}{c|}{(A1) $10^{-10}$	} 	&\multicolumn{2}{c|}{(A2) Endpoints} 	&\multicolumn{2}{c|}{(A3) Global} &\multicolumn{2}{c|}{(B) Sparse} &\multicolumn{2}{c}{(C) PCA}\\
	$N$	&$\sigma$		&mean 	&std		&mean  &std 	&mean		&std		&mean  &std 	&mean	&std &mean &std \\
	\hline
	\multirow{8}{*}{4}
% 	Done on HPC, Jap_vowels_Compare_forNN_all2.m
%	28 Nov, 2016
&0.00  &{\bf \color{red}95.99} &0.59      &{\bf \color{red}95.90} &0.59     	&94.92 &1.01      &90.27 &0.71     &{\bf \color{red}95.89} &0.67      	&95.33 &0.93 \\
&0.01  &{\bf \color{red}95.93} &0.58      &{\bf \color{red}95.79} &0.60     	&94.66 &1.06      &90.39 &0.91     &{\bf \color{red}95.81} &0.65      	&95.32 &0.86 \\
&0.05  &{\bf \color{red}95.31} &2.13      &{\bf \color{red}95.03} &2.66     	&89.75 &1.93      &90.04 &0.88     &{\bf \color{red}95.20} &2.59      	&95.04 &0.89 \\
&0.10  &93.17 &5.13      &92.53 &5.90     	&74.59 &3.26      &88.85 &0.97     &92.97 &5.76      	&{\bf \color{red}93.75} &1.23 \\
&0.15  &91.16 &5.88      &90.22 &6.51     	&58.06 &3.68      &87.04 &1.19     &90.90 &6.47      	&{\bf \color{red}91.66} &1.50 \\
&0.20  &86.69 &6.82      &85.55 &8.16     	&46.04 &3.23      &84.62 &1.46     &86.41 &8.07      	&{\bf \color{red}88.25} &2.16 \\
&0.25  &80.70 &9.13      &78.75 &12.00     	&38.62 &3.08      &81.70 &1.62     &79.86 &12.23    	&{\bf \color{red}83.58} &2.84\\ 
&0.30  &75.23 &9.20      &73.63 &11.04     	&32.64 &2.57      &77.36 &2.17     &74.76 &11.47      	&{\bf \color{red}77.55} &4.02 \\
%&0.40  &65.84 &7.85      &64.75 &8.00     	&25.86 &2.30      &{\bf \color{red}69.37} &2.25     &66.02 &8.29      	&66.01 &5.10 \\
	\hline
	\multirow{8}{*}{10}
%	Done on HPC, Jap_vowels_Compare_forNN_all2.m
%	28 Nov 2016
&0.00  &{\bf \color{red}96.12} &0.56      &{\bf \color{red}95.93} &0.57     &{\bf \color{red}96.09} &0.93      &91.26 &1.17     &{\bf \color{red}96.09} &0.65      &94.71 &0.93 \\
&0.01  &{\bf \color{red}96.28} &0.55      &{\bf \color{red}96.06} &0.61     &{\bf \color{red}96.09} &0.92      &91.09 &1.16     &{\bf \color{red}96.15} &0.60      &94.65 &1.11 \\
&0.05  &{\bf \color{red}95.37} &2.00      &94.34 &7.02     &91.48 &1.64      &90.55 &1.20     &94.58 &6.98          &94.02 &1.19 \\
&0.10  &{\bf \color{red}93.09} &4.20      &91.66 &8.38     &77.51 &3.15      &89.31 &1.30     &92.22 &8.15          &92.44 &1.55 \\
&0.15  &{\bf \color{red}88.34} &8.16      &86.96 &9.91     &60.50 &4.18      &86.94 &1.57     &87.62 &10.20        &{\bf \color{red}88.34} &2.13 \\
&0.20  &83.08 &8.84      &81.18 &10.76     &49.59 &4.21      &{\bf \color{red}83.99} &1.78     &82.18 &11.08      &81.16 &3.09 \\
&0.25  &78.62 &9.57      &76.85 &11.75     &39.88 &3.43      &{\bf \color{red}80.33} &2.12     &77.74 &12.25      &71.16 &4.03 \\
&0.30  &72.20 &9.67      &70.19 &11.35     &34.19 &2.96      &{\bf \color{red}76.75} &2.39     &71.65 &11.73      &68.67 &4.57 \\
%&0.40  &59.28 &8.21      &57.74 &9.41       &26.60 &2.76      &{\bf \color{red}67.44} &2.52     &58.83 &10.18      &61.41 &4.38 \\
	\hline
	\multirow{8}{*}{15}
%	Done on HPC, Jap_vowels_Compare_forNN_all2.m
%	28 Nov 2016
&0.00  &96.34 &0.58      	&96.15 &0.59     	&{\bf \color{red}96.74} &0.76      &91.66 &1.32     &96.29 &0.59      	&94.31 &1.07 \\
&0.01  &{\bf \color{red}96.39} &0.59      	&96.08 &0.84     	&{\bf \color{red}96.27} &0.87      &91.58 &1.43     &{\bf \color{red}96.25} &0.95      	&93.57 &1.33 \\
&0.05  &{\bf \color{red}95.76} &1.01      	&95.24 &1.84     	&92.41 &1.51      &90.97 &1.38     &{\bf \color{red}95.58} &1.76      	&92.99 &1.46 \\
&0.10  &{\bf \color{red}93.39} &3.48      	&92.14 &6.16     	&79.93 &3.52      &89.67 &1.30     &92.69 &6.25      	&90.80 &1.60 \\
&0.15  &{\bf \color{red}87.44} &8.38      	&85.01 &12.08     	&63.95 &4.13      &87.18 &1.56     &85.94 &12.30      	&85.61 &2.37\\ 
&0.20  &82.34 &8.31      	&80.46 &9.82     	&51.04 &3.69      &{\bf \color{red}83.73} &1.67     &81.56 &9.99      	&80.68 &3.06 \\
&0.25  &75.65 &10.53      	&73.59 &12.69     	&41.84 &3.83      &{\bf \color{red}80.23} &2.14     &74.58 &12.84      	&76.09 &3.71 \\
&0.30  &67.10 &11.30      	&65.41 &12.14     	&34.39 &3.51      &{\bf \color{red}75.24} &2.87     &66.82 &12.43      	&72.64 &3.36 \\
%&0.40  &54.95 &9.86      	&53.39 &10.97     	&27.26 &3.07      &64.80 &3.70     &54.61 &11.36      	&{\bf \color{red}66.58} &3.91 \\
	\hline
	\multirow{8}{*}{20}
%	Done on HPC, Jap_vowels_Compare_forNN_all2.m
%	28 Nov 2016
&0.00  &96.42 &0.56      &96.25 &0.56     &{\bf \color{red}97.00} &0.75      &92.16 &1.20     &96.37 &0.59      &94.01 &1.17 \\
&0.01  &96.39 &0.53      &96.15 &0.65     &{\bf \color{red}96.84} &0.77      &92.31 &1.18     &96.33 &0.62      &93.61 &1.14 \\
&0.05  &{\bf \color{red}95.50} &1.40      &93.81 &6.08     &93.16 &1.54      &91.68 &1.28     &94.19 &6.04      &93.21 &1.29 \\
&0.10  &{\bf \color{red}93.31} &3.52      &91.65 &6.52     &81.37 &3.26      &90.16 &1.45     &92.39 &6.31      &90.73 &1.76 \\
&0.15  &{\bf \color{red}89.16} &5.46      &86.87 &9.01     &65.75 &4.75      &87.48 &1.60     &87.92 &9.12      &87.38 &2.29 \\
&0.20  &80.66 &9.88      &77.23 &14.46     &52.01 &3.77      &{\bf \color{red}83.67} &2.22     &78.47 &14.61      &{\bf \color{red}83.61} &2.46 \\
&0.25  &75.87 &7.83      &73.45 &11.57     &42.63 &4.74      &{\bf \color{red}79.15} &2.62     &74.94 &11.28      &{\bf \color{red}78.99} &3.14 \\
&0.30  &65.17 &11.29      &62.45 &13.18     &35.50 &3.34      &73.87 &3.14     &63.65 &13.80      &{\bf \color{red}75.86} &3.36 \\
%&0.40  &51.24 &9.93      &49.17 &11.61     &26.75 &2.98      &63.24 &4.18     &50.13 &12.00      &{\bf \color{red}71.15} &3.24 \\
	\hline
	\multirow{8}{*}{50}
%TsAccLin04 TsAccLin10 TsAccLinEnd TsAccSingle TsAccDan TsAccPCA
% N = 50 done on Nov 27, 2016 using Jap_vowels_Compare_forNN_all.m
%	Saved as: Results_Compare_all_N50.mat
% tolerances: 0.975, 0.95,   0.95    0.95,   0.80,   0.80,   0.65,   0.30,   0.30;
&0.00  &96.58 &0.52      &96.46 &0.55     	&{\bf \color{red}97.74} &0.63      &94.65 &0.95     &96.60 &0.56      &95.12 &0.74 \\
&0.01  &96.52 &0.53      &96.16 &0.87     	&{\bf \color{red}97.58} &0.77      &94.51 &0.86     &96.34 &0.82      &95.21 &0.91 \\
&0.05  &{\bf \color{red}95.76} &1.15      &93.84 &4.42     	&95.16 &1.25      &94.16 &0.97     &94.15 &4.25      &94.88 &0.82 \\
&0.10  &92.46 &3.29      &85.76 &14.92     	&84.40 &3.27      &92.53 &1.27     &86.61 &14.83      &{\bf \color{red} 93.72} &1.12\\
&0.15  &84.89 &5.35      &75.20 &17.20     	&67.75 &4.43      &89.21 &1.42     &76.49 &17.46      &{\bf \color{red} 91.98} &1.21 \\
&0.20  &73.85 &8.90      &66.38 &15.80     	&52.45 &4.37      &84.25 &2.23     &68.28 &15.96      &{\bf \color{red} 89.35} &1.60 \\
&0.25  &63.04 &8.85      &57.11 &13.61     	&40.77 &4.78      &77.20 &3.52     &58.59 &14.04      &{\bf \color{red} 85.88} &1.63 \\
&0.30  &52.10 &8.44      &46.29 &12.36     	&33.79 &3.84      &70.02 &3.96     &48.08 &12.79      &{\bf \color{red} 82.58} &2.18 \\
%&0.40  &36.75 &8.25      &33.32 &10.28     	&24.53 &3.34      &55.23 &5.09     &34.50 &10.67      &{\bf \color{red} 77.95} &2.25 \\
	\hline
	\multirow{8}{*}{75}
% 	N = 75 done on Nov 27 using Jap_vowels_Compare_forNN_all.m
%	Saved as: Results_Compare_all_N75.mat
%	    tolerances: 0.9750    0.9500    0.9500    0.9500    0.9500    0.8000    0.7500    0.6500    0.3000
&0.00  &96.75 &0.56      &96.49 &0.59     &{\bf \color{red}97.66} &0.76      &95.27 &0.78     &96.68 &0.62      &95.43 &0.75 \\
&0.01  &96.60 &0.55      &96.24 &0.67     &{\bf \color{red}97.62} &0.66      &95.22 &0.80     &96.45 &0.78      &95.65 &0.72 \\
&0.05  &{\bf \color{red}95.98} &0.87      &91.74 &9.68     &94.69 &1.39      &94.72 &0.87     &92.36 &9.47      &95.49 &0.81 \\
&0.10  &91.84 &2.49      &81.51 &14.44     &82.55 &3.97      &93.36 &1.04     &83.10 &14.31      &{\bf \color{red}94.62} &0.85 \\
&0.15  &82.97 &4.80      &68.99 &17.22     &66.42 &5.33      &89.93 &1.66     &70.52 &17.73      &{\bf \color{red}93.05} &1.14 \\
&0.20  &70.09 &6.25      &56.91 &16.28     &49.61 &4.67      &83.53 &2.58     &58.97 &16.80      &{\bf \color{red}90.93} &1.38 \\
&0.25  &58.25 &5.78      &47.91 &12.66     &38.93 &3.96      &75.49 &3.62     &49.88 &13.09      &{\bf \color{red}87.27} &1.67 \\
&0.30  &45.51 &6.98      &37.06 &11.37     &31.02 &4.05      &66.92 &4.36     &38.80 &11.65      &{\bf \color{red}84.63} &1.83\\ 
%&0.40  &30.32 &5.22      &25.36 &7.31     &21.59 &3.15      &50.89 &5.68     &26.07 &7.59      &{\bf \color{red}80.46} &2.01 \\
	\hline
	\multirow{8}{*}{100}
% 	N = 75 done on Nov 27 using Jap_vowels_Compare_forNN_all.m
%	Saved as: Results_Compare_all_N100.mat
%	    tolerances: 0.975, 0.95,   0.95,   0.95,   0.95,   0.80,   0.75,   0.65,   0.45
&0.00  &96.69 &0.57      &96.32 &0.63     	&{\bf \color{red}97.33} &0.73      &95.55 &0.70     &96.54 &0.69      &95.63 &0.62 \\
&0.01  &96.64 &0.60      &96.00 &0.83     	&{\bf \color{red}97.24} &0.83      &95.55 &0.74     &96.38 &0.82      &95.91 &0.61 \\
&0.05  &{\bf \color{red}95.82} &0.78      &87.79 &11.07     	&93.88 &1.47      &95.21 &0.87     &89.04 &10.50      &{\bf \color{red}95.59} &0.72 \\
&0.10  &91.55 &2.18      &72.86 &17.72     	&80.51 &4.35      &94.03 &0.94     &74.40 &18.00      	&{\bf \color{red}94.84} &0.83 \\
&0.15  &80.22 &4.36      &61.16 &16.07     	&62.60 &5.49      &90.17 &1.76     &63.32 &16.18      	&{\bf \color{red}93.51} &1.06 \\
&0.20  &66.44 &5.15      &46.40 &14.00     	&46.78 &4.83      &83.87 &2.41     &48.01 &14.68      	&{\bf \color{red}91.51} &1.35 \\
&0.25  &51.70 &5.29      &35.58 &11.05     	&34.83 &4.37      &74.86 &3.69     &37.12 &12.17      	&{\bf \color{red}88.18} &1.44\\ 
&0.30  &39.20 &5.22      &28.47 &8.47     	&28.30 &3.92      &64.53 &5.07     &29.91 &9.18      	&{\bf \color{red}85.24} &1.79 \\
%&0.40  &25.49 &4.39      &20.46 &5.11     	&19.94 &3.15      &47.58 &5.59     &21.32 &5.18      	&{\bf \color{red}80.64} &1.98 \\
\hline
\end{tabular}
\caption{Means and standard deviations of the percent classification accuracy over 100 simulations on the noisy Japanese vowel test dataset using various classification methods.  For each reservoir size $N$ and noise level $\sigma$, the best results within $0.25\%$ are highlighted in red boldface font.}
\label{tab:jv accuracy}
\end{table*}

 Figures~\ref{fig:JPAcc} and~\ref{fig:JV Accuracy} display more detailed classification accuracy results.   Figure~\ref{fig:JPAcc} fixes the reservoir size at $N=75$, and displays the mean and standard deviation of the test accuracy over all 100 simulations at each noise level $\sigma$.  Figure~\ref{fig:JV Accuracy}  fixes the noise level at $\sigma = 0.2$, and displays the mean test accuracy at each reservoir size.  The right-hand plot of Figure~\ref{fig:JV Accuracy} displays the `boxed' area of the left-hand plot, and includes standard deviation information as vertical bars.  The figures show that the low-rank approximation method C performs well, tending to have the highest classification accuracy and lowest variation of results. Notice that most of the methods have similar classification accuracy for small reservoir sizes as in the right-hand plot of Figure~\ref{fig:JV Accuracy}, but the low-rank method C and single global weights method A3 have much lower variation than the others.

\section{Conclusion}\label{sec:conclusion}

In this study, several approaches for performing classification with ESNs were compared.  The numerical experiments suggest that the classification accuracy depends strongly on the characteristics of the data, the classification method used, and the design of the ESN.   It was observed that in both synthetic and real-world datasets with low noise corruption, employing an output method using linear readout weights trained on the final temporal state of a reservoir, as in Equation~\eqref{eq:end}, with moderate number of nodes produces the most accurate results.  On the other hand, if the data are noisy then the findings suggest that one should design a reservoir with a larger number of nodes and use the classification method of finding a low-rank approximation of the reservoir states, as in Equation~\eqref{eq:pca}.  In general, we did not observe evidence to support using the classification method using sparse linear readout weights in practice.  Future directions include investigating on additional datasets and with other reservoir types and output methods, including time delay reservoirs, multilayer perceptrons, and a global variation of the low-rank approximation method, rather than the pointwise approach studied here.

\subsection*{Acknowledgments}
\noindent This research was partially supported by Air Force Office of Scientific Research [LRIR:15RICOR122]. \\

\noindent Approved for public release by USAF 88 ABW on 18 NOV 2016.  Case Number: 88ABW--2016--5889.   Any opinions, findings and conclusions or recommendations expressed in this material are those of the author and do not necessarily reflect the view of the United States Air Force.

%%%%%%%%%%%%%%%%%%%%%%%%%%%%%%%%%%%%%%%%%%
\bibliographystyle{elsarticle-num}
\bibliography{mybibfile}

\end{document}